\documentclass{article}

\usepackage{spconf,amsmath,graphicx}
\usepackage{todonotes}
\usepackage{lipsum} 
\usepackage{amsmath} 
\usepackage{booktabs}
\usepackage{multirow}
\usepackage{caption}
\usepackage{subcaption}
\usepackage{url}

\newcommand{\methodname}{CoReSeg} 
\newcommand{\currprop}{\textwidth}

\title{Conditional Reconstruction for Open-set Semantic Segmentation}
\name{Ian Nunes$^{\dagger}$, Matheus B. Pereira$^{\star}$, Hugo Oliveira$^{\diamond}$, Jefersson A. dos Santos$^{\star}$, Marcus Poggi$^{\dagger}$\thanks{This research was partially financed by the Coordenação de Aperfeiçoamento de Pessoal de Nível Superior (CAPES), Fundação de Amparo à Pesquisa do Estado de Minas Gerais (FAPEMIG), Fundação de Amparo à Pesquisa do Estado de São Paulo (FAPESP -- grant \#2020/06744-5), Agence Nationale de la Recherche (ANR) and the Serrapilheira Institute (grant R-2011-37776). The authors would also like to thank NVIDIA for the donation of the GPUs used in this research.} \thanks{ \textcopyright 2022 IEEE. Personal use of this material is permitted. Permission from IEEE must be obtained for all other uses, in any current or future media, including reprinting/republishing this material for advertising or promotional purposes, creating new collective works, for resale or redistribution to servers or lists, or reuse of any copyrighted component of this work in other works.}}
\address{
$^{\dagger}$Pontifícia Universidade Católica do Rio de Janeiro, Rio de Janeiro, Brazil \\
$^{\star}$Universidade Federal de Minas Gerais, Belo Horizonte, Brazil \\
$^{\diamond}$Universidade de São Paulo, São Paulo, Brazil \\
\{inunes, poggi\}@inf.puc-rio.br, \{matheuspereira,  jefersson\}@dcc.ufmg.br, oliveirahugo@ime.usp.br
}

\begin{document}
%
\maketitle
\begin{abstract}
Open set segmentation is a relatively new and unexplored task, with just a handful of methods proposed to model such tasks. 
We propose a novel method called \methodname{} that tackles the issue using class conditional reconstruction of the input images according to their pixelwise mask.
Our method conditions each input pixel to all known classes, expecting higher errors for pixels of unknown classes.
It was observed that the proposed method produces better semantic consistency in its predictions, resulting in cleaner segmentation maps that better fit object boundaries.
\methodname{} outperforms state-of-the-art methods on the Vaihingen and Potsdam ISPRS datasets, while also being competitive on the Houston 2018 IEEE GRSS Data Fusion dataset. Official implementation for \methodname{} is available at: \url{https://github.com/iannunes/CoReSeg}.

\end{abstract}
\section{Introduction}
\label{sec:intro}

\renewcommand{\currprop}{0.95\columnwidth}
\begin{figure}[!t]
    \centering
    \includegraphics[width=\currprop]{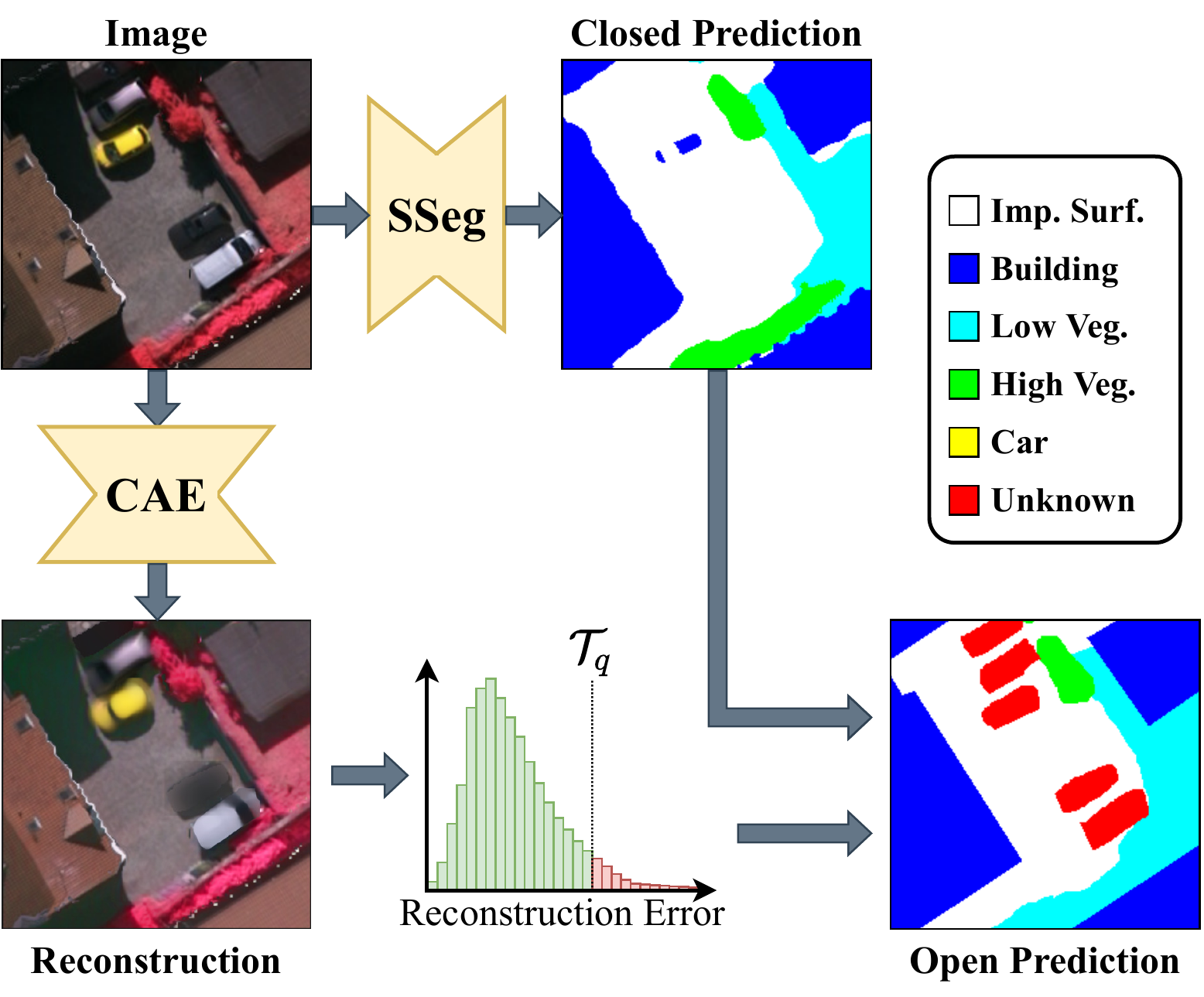}
    \caption{Overview of \methodname{} for OS3. An SSeg model and a Conditional AutoEncoder (CAE) are trained for closed set segmentation and conditional reconstruction of the input image, respectively. Reconstruction errors are then thresholded according to a quantile $q$ to determine unknown pixels.}
    \label{fig:intro}
\end{figure}

In the last decade many computer vision tasks such as object classification \cite{krizhevsky2012imagenet}, detection \cite{ren2015faster} and semantic segmentation \cite{long2015fully,ronneberger2015} have greatly improved their performance. Semantic Segmentation is a fundamental task in computer vision field, and its objective is to assign a label to each pixel of the image \cite{YUAN2021114417}. Traditional Semantic Segmentation (SSeg) tasks use closed set models, since the produced segmentation models are conceived to learn and predict the same set of classes. However, assuming an open set of classes is more applicable in real-world scenarios, where classes not seen during training appear on the deploy phase. In such cases, the method must be able to identify the pixels of unseen or unknown classes while still correctly segmenting known classes \cite{geng2021,silva2020,oliveira2021}, which is the definition of Open Set Semantic Segmentation (OS3).

OS3 is inherently harder than open set classification due to its dense labeling nature. This can explain the gap in the literature for OS3 methods, with only a handful of works tackling the issue \cite{cui2020}. According to Silva \textit{et al.} \cite{silva2020} a dataset with all possible classes known in advance will hardly be found in practice, especially when dealing with remote sensing datasets. Silva \textit{et al.} \cite{silva2020} adapted the well-known OpenMax algorithm \cite{bendale2016} for patchwise OS3, resulting in the OpenPixel method. OpenPixel, however, has a major drawback of being extremely inefficient during both training and testing. A nonparametric statistical adaptive OS3 method was proposed by Cui \textit{et al.} \cite{cui2020}. Employing the Mann-Whitney U test on a closed set segmentation output to determine the existence of unknown classes in each image and uses an adaptive threshold that defines which pixels are defined as unknown.

The first fully convolutional architectures for OS3 were proposed by Oliveira \textit{et al.} \cite{oliveira2021}. Open Fully Convolutional Network (OpenFCN) is a fully convolutional extension of OpenPixel \cite{silva2020} for dense labeling tasks. Open Principal Component Scoring (OpenPCS) \cite{oliveira2021} uses the intermediate feature activations from closed set networks to fit gaussian distributions in a low-dimensional projection, obtaining a likelihood score for OOD pixel detection.

The performance of all previously mentioned OS3 methods vary considerably across datasets and distinct hidden class scenarios. Thus, robustness is still lacking in OS3 algorithms, with qualitative results varying widely across tasks and poor semantic consistency along neighboring pixels in OS3 predictions. With these limitations in mind, we propose a novel end-to-end fully convolutional OS3 method named Conditional Reconstruction for Open Set Segmentation (\methodname). Figure~\ref{fig:intro} shows an overview of the information flow in \methodname{}, with reconstruction and closed set segmentation branches merging into one single open set prediction. We describe this novel development in Section~\ref{sec:method} and present experiments and our conclusions along Sections~\ref{sec:setup},~\ref{sec:results} and~\ref{sec:conclusion}. Our experiments show that \methodname{} is capable of yielding OS3 predictions with superior semantic consistency, overall robustness and better average AUROC when compared to the state-of-the-art. 
\section{Proposed Method: \methodname}
\label{sec:method}




\renewcommand{\currprop}{0.95\columnwidth}

\begin{figure}[!t]
    \centering
    \begin{subfigure}[b]{\columnwidth}
        \centering
        \includegraphics[width=\currprop]{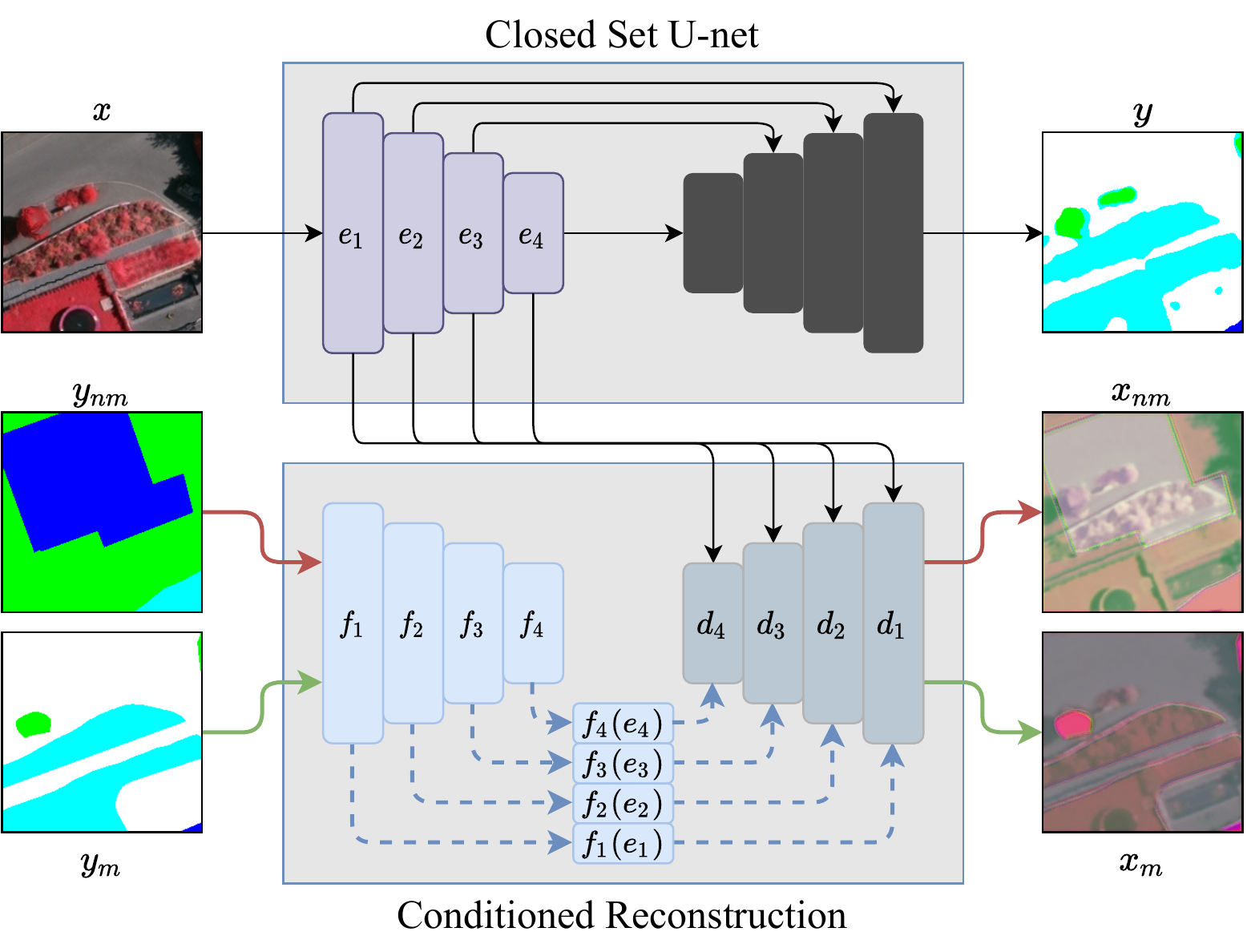}
        \caption{}
        \label{fig:training}
    \end{subfigure}
    \hfill
    \begin{subfigure}[b]{\columnwidth}
        \centering
        \includegraphics[clip, trim=0in 0in 0.45in 0in, width=\currprop]{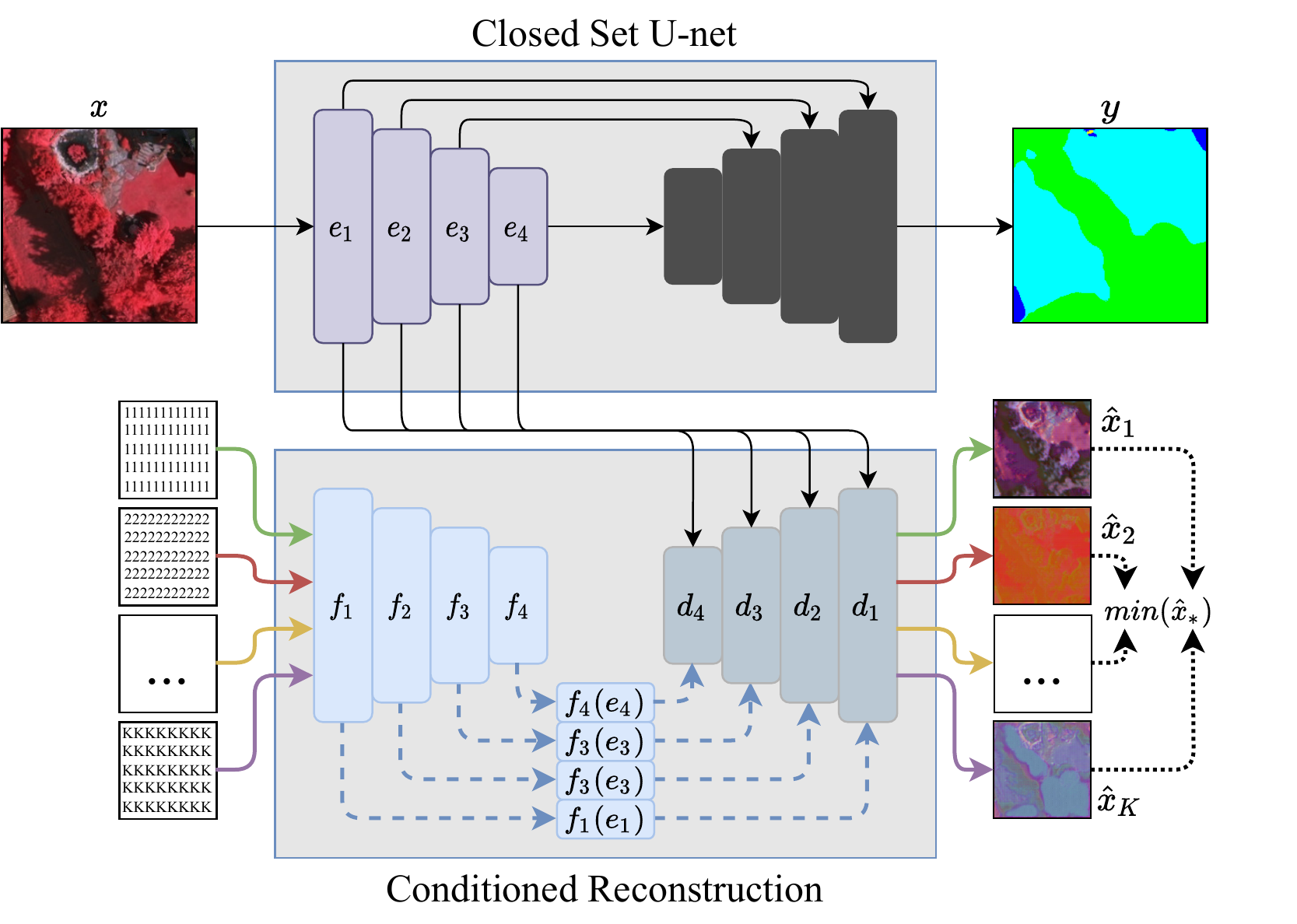}
        \caption{}
        \label{fig:testing}
    \end{subfigure}
    \caption{Open Set Training (a) and Deploy (b) schematics for \methodname. $e_i$ denotes a layer block on the Closed Set Encoder, $d_i$ denotes a block on the reconstruction decoder, and $f_i$ denotes a simplified conditioning layer that has two encoders $\beta$ and $\gamma$ corresponding the parameters of a pixelwise FiLM conditioning \cite{perez2018film} operation. Introducing the match/non match conditional reconstruction training procedure in (a), and in (b) evaluating the model by conditioning the input image with each of the known classes to find the pixelwise minimum loss.
    }
    \label{fig:coreseg}
\end{figure}
     


\methodname{} employs a pre-trained closed set neural network to generate a latent representation of the input image, which is then sent to a reconstruction decoder. This decoder also receives conditioning information, whose objective is to guide the reconstruction of the input image from its latent features conditioned to the desired class (or classes). An overview of this process is illustrated in Figure~\ref{fig:training}.

Our method is inspired in some design choices proposed by Oza \textit{et al.} \cite{oza_patel_2019}, wherein the conditioning concerns the entire image, while 
for \methodname{} the conditioning is performed in a pixel-wise manner. 
The main idea of the conditioning procedure is that objects from known classes will be better reconstructed when their pixels are conditioned to the correct class, while unknown objects will present poor reconstructions, since there is no correct conditioning to these pixels.

\methodname{} has 3 sequential stages to be further explained in the following paragraphs: 1) Closed Set Training, 2) Conditional Reconstruction, and 3) Open Set Pixel Recognition.

\noindent\textbf{Closed Set Training:} \methodname{} requires a pre-trained closed set segmentation model to generate latent features that will be later used as input to the reconstruction decoder. For this reason, the first necessary step is to train a U-net \cite{ronneberger2015} in a traditional closed set scenario, where only the known classes are learned. We call the encoder of this U-net the Closed Set Encoder, and its output is the desired latent representation for the conditional reconstruction training to be performed later. The resulting model will be used in both \textit{Open Set Training} and \textit{Deploy} phases. Also, the weights of this closed set U-net will not change after its initial training, which means that the semantic segmentation layers are frozen during the training of the rest of \methodname's framework.

\noindent\textbf{Conditional Reconstruction:} Conditional Reconstruction training aims at reconstructing the image that serves as input to the closed set semantic segmentation block of the framework from its latent representation. The reconstruction is conducted by a conditioning input, which, in the case of a semantic segmentation task, is comprised by a mask providing a class for each pixel. The conditional reconstruction block of the framework can be viewed as an AutoEncoder where the conditioning layers are the encoder and the reconstruction layers are the decoder. Figure~\ref{fig:training} shows how these different parts of the training are connected to each other.

In order to train the model to reconstruct the image taking into account the conditioning input, \methodname{} uses for each input image 2 different masks to condition the reconstruction. The first mask is the one correctly labeled for the image (match mask, $y_{m}$). The second one is the label from a different image (non-match mask, $y_{nm}$), which means that it incorrectly conditions the pixels from the input image. The use of a non-match mask is important to make sure the network is actually learning to condition the input while not simply reconstructing the image from the latent representation.

We employ the L1 loss in the reconstruction step. The final reconstruction loss $\mathcal{L}$ is computed as follows: 
\begin{equation}
   \mathcal{L} = L1(x, \hat{x}_{m}) + \alpha * L1(x, \hat{x}_{nm}),
   \label{eq:loss}
\end{equation}
where $x$ is the input image, $\hat{x}_{m}$ and $\hat{x}_{nm}$ are the reconstructions conditioned on $y_{m}$ and $y_{nm}$ respectively, while $\alpha$ weights the importance of each term.

Aiming to enforce the conditioning, the encoder from the conditional reconstruction block applies a transformation to the intermediate features from the frozen Closed Set Encoder layers ($e_i$). The result of this transformation is then used as input on the corresponding layer of the reconstruction decoder ($d_i$). In Figure~\ref{fig:training}, this process is represented by the $f_i(e_i)$ blocks and it is performed for both match and non-match conditioning masks. The transformation responsible for the conditioning is the FiLM method proposed by Perez \textit{et al.} \cite{perez2018film} extended to work in a pixelwise problem. More specifically, in order to use the pixelwise FiLM, the conditional reconstruction decoder is composed by two auxiliary encoders: $\beta$ and $\gamma$. Both encoders have the same shape as $e_i$ and $d_i$. In order to apply the transformation, we perform the following operation $\gamma_i \odot e_i + \beta_i$, where $\beta_i$ and $\gamma_i$ are the $i^{th}$ blocks of the conditional reconstruction encoder, and $e_i$ is the output of the $i^{th}$ block from the Closed Set Encoder. This procedure allows us to perform pixelwise FiLM conditioning on $e_i$.

The reconstruction decoder uses as a main input the latent representation of the input image from the Closed Set Encoder. Furthermore, each layer of the reconstruction decoder receives two additional inputs concatenated to the previous layer activation: (i) the corresponding FiLM transformation ($f_i$) from the conditional reconstruction encoder, and (ii) the raw feature maps from the corresponding Closed Set Encoder ($e_i$). These concatenations can also be viewed in Figure~\ref{fig:training}.



\noindent\textbf{Open Set Pixel Recognition.} During deploy -- shown in Figure~\ref{fig:testing} -- we cannot provide match and non-match masks for the conditional encoder, as the labels for these samples are not available. So, to define which pixels are known and unknown \methodname{} tries to condition every pixel for each known class. 
Then, for each pixel the minimum loss for $k \in \{1, 2, \dots, K\}$ is computed and selected -- where $\{1, 2, \dots, K\}$ is the set of known classes. Pixels that were conditioned to the right class yield a small minimum error, while unknown pixels result in higher loss values for each one of the reconstructions, since none of them match the right expected class. At last, a threshold operation defines which pixels are known and unknown. We use error quantiles to set thresholds and find the best performance for the model. If the minimum reconstruction loss of a pixel is below the threshold, its class is deemed as known and set to the closed set predicted output, and otherwise it is set as unknown.


\section{Experimental Setup}
\label{sec:setup}


Three datasets were selected to evaluate the proposed method: Vaihingen\footnote{https:\slash\slash www2.isprs.org\slash commissions\slash comm2\slash wg4\slash benchmark\slash \label{foot:isprs}} and Potsdam\textsuperscript{1} datasets; and the Houston GRSS 2018 Data Fusion dataset \cite{GRSS}. As the selected datasets for this work are intended for closed set segmentation, we emulate open set environments using the Leave One Class Out (LOCO) protocol \cite{oliveira2021}. 
As the final model for the open set pixel recognition process, we choose the model with the best AUROC result calculated on the validation set.







\renewcommand{\currprop}{0.7\textwidth}
\begin{figure*}[!t]
    \centering
    \includegraphics[width=\currprop]{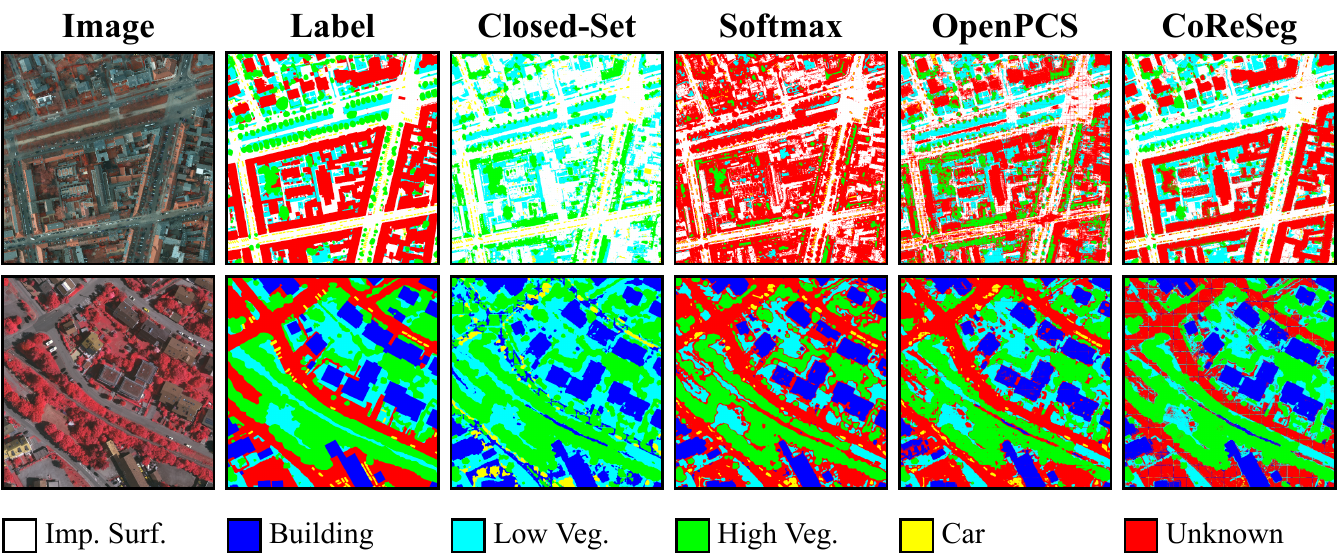}
    \caption{Visual results for Potsdam (UUC Impervious Surfaces, top row) and Vaihingen (UUC Building, bottom row) using the LOCO protocol for the closed set model, Softmax with thresholding, OpenPCS and \methodname.}
    \label{fig:qualitative}
\end{figure*}

\begin{table*}[]
    \centering
    \renewcommand{\arraystretch}{0.7}
    \begin{tabular}{cc|cccccc|cccccc}
        \toprule
        \multicolumn{1}{c}{\multirow{3}{*}{\textbf{Method}}} & \multicolumn{1}{c}{\multirow{3}{*}{\textbf{Backbone}}} & \multicolumn{6}{c}{\textbf{Vaihingen}} & \multicolumn{6}{c}{\textbf{Potsdam}} \\ 
        \cmidrule(l){3-8} \cmidrule(l){9-14} 
        \multicolumn{1}{c}{} & \multicolumn{1}{c}{} & \multicolumn{5}{c}{\textbf{UUCs}} & \multicolumn{1}{c}{\multirow{2}{*}{\textbf{Avg.}}} & \multicolumn{5}{c}{\textbf{UUCs}} & \multicolumn{1}{c}{\multirow{2}{*}{\textbf{Avg.}}} \\ \cmidrule(lr){3-7} \cmidrule(lr){9-13}
        \multicolumn{1}{c}{} & \multicolumn{1}{c}{} & \textbf{0} & \textbf{1} & \textbf{2} & \textbf{3} & \textbf{4} & \multicolumn{1}{c}{} & \textbf{0} & \textbf{1} & \textbf{2} & \textbf{3} & \textbf{4} & \multicolumn{1}{c}{} \\ \midrule
        Softmax & DN-121 & .78 & .63 & .73 & .67 & .58 & $.678 \pm .08$ & .66 & .49 & .53 & .64 & .76 & $.616 \pm .11$ \\
        Softmax & WRN-50 & .80 & .64 & .75 & .63 & .58 & $.680 \pm .09$ & .68 & .53 & .60 & .62 & .66 & $.618 \pm .06$\\
        Softmax & U-net & .86 & .63 & .76 & .69 & .62 & $.708 \pm .10$ & .75 & .50 & .74 & \textbf{.65} & .68 & $.664 \pm .10$\\
        OpenPCS & DN-121 & .84 & \textbf{.94} & .70 & .81 & .65 & $.788 \pm .12$ & .83 & .87 & .28 & .54 & .77 & $.658 \pm .25$\\
        OpenPCS & WRN-50 & .87 & \textbf{.94} & \textbf{.77} & .63 & \textbf{.87} & $.816 \pm .12$ & .66 & .86 & \textbf{.71} & .57 & \textbf{.86} & $.732 \pm .13$\\
        OpenPCS & U-net & \textbf{.90} & .80 & .51 & .84 & .54 & $.717 \pm .18$ & \textbf{.88} & .86 & .42 & .53 & .84 & $.705 \pm .21$\\ \midrule
        \methodname{} (ours) & U-net & .88 & .93 & .71 & \textbf{.87} & \textbf{.87} & $\textbf{.854} \pm \textbf{.08}$ & .82 & \textbf{.90} & .70 & .63 & .77 & $\textbf{.764} \pm \textbf{.07}$ \\ \bottomrule
    \end{tabular}
    \caption{Classwise AUROC and average (\textit{Avg.}) AUROC in Vaihingen and Potsdam. Numbered columns stand for: 1 -- Impervious Surfaces, 2 -- Building, 3 -- Low Vegetation, 4 -- High Vegetation and 5 -- Car.}
    \label{tab:results_vaihingen_potsdam}
\end{table*}

\begin{table}[!h]
    \centering
    \renewcommand{\arraystretch}{0.7}
    \begin{tabular}{@{}ccccc@{}}
        \toprule
        \multicolumn{1}{c}{\multirow{2}{*}{\textbf{Method}}} & \multicolumn{1}{c}{\multirow{2}{*}{\textbf{Backbone}}} & \multicolumn{2}{c}{\textbf{UUCs}} & \multicolumn{1}{c}{\multirow{2}{*}{\textbf{Avg.}}} \\ \cmidrule(lr){3-4}
        \multicolumn{1}{c}{} & \multicolumn{1}{c}{} & \textit{\textbf{Veg.}} & \textit{\textbf{Build.}} & \multicolumn{1}{c}{} \\ \midrule
        Softmax & DN-121 & .61 & .66 & $.635 \pm .04$ \\ 
        Softmax & U-net & .58 & .75 & $.666 \pm .12$ \\ 
        OpenPCS & DN-121 & .81 & .76 & $.785 \pm .04$ \\ 
        OpenPCS & U-net & .65 & .67 & $.660 \pm .01$ \\ 
        OpenIPCS & DN-121 & \textbf{.90} & \textbf{.88} & $\textbf{.890} \pm \textbf{.01}$ \\ 
        OpenIPCS & U-net & .69 & .68 & $.683 \pm .01$\\ \midrule
        \methodname{} (ours) & U-net & .82 & .74 & $.780 \pm .05$ \\ \bottomrule
    \end{tabular}
    \caption{AUROC of Vegetation (\textit{Veg.}) and Building (\textit{Build.}) UUCs, as well as the average (\textit{Avg.}) AUROC in Houston.}
    \label{tab:grss_results}
\end{table}

\noindent\textbf{Vaihingen and Potsdam:} 
Both datasets have images with 4 spectral channels IR-R-G-B paired with the normalized digital surface model (nDSM). The semantic maps contain 6 classes: impervious surfaces, building, low vegetation, high vegetation, car and miscellaneous. All experiments were conducted using the IR-R-G-nDSM channels as input, leaving the Blue channel out. We separated images from the original training/test set used by OpenPCS \cite{oliveira2021} to serve as a validation set: area 23 for Vaihingen; and areas 3\_11 and 6\_14 for Potsdam. For both datasets we tested 5 different scenarios in which one class was selected as UUC for each case.



\noindent\textbf{Houston:} For this work only the RGB spectral channels and DSM image were used. The image fusion procedure we employed was the same as described in OpenPCS \cite{oliveira2021}. 
On Houston we tested 2 different scenarios: 1) \textit{Vegetation}, wherein the UUC is composed by healthy grass, stressed grass, 
evergreen trees, and deciduous trees; and 2) \textit{Building}, in which the UUCs are residential and non-residential buildings.

\section{Results and Discussion}
\label{sec:results}



As presented in Table~\ref{tab:results_vaihingen_potsdam}, for Vaihingen and Potsdam, \methodname{} attained the best average AUROC result. \methodname{} reached a higher and more uniform performance between the different tested UUCs. As it can be seen in Figure~\ref{fig:qualitative} the OS3 obtained using \methodname{} resulted in a much cleaner result than other methods with similar or even higher AUROC performance. The final segmentation is more semantically consistent and looks closer to ground truth.


As shown in Table~\ref{tab:results_vaihingen_potsdam}, until this work the state-of-the-art performance for both Vaihingen and Potsdam was achieved by OpenPCS with WRN-50. \methodname{} achieves better average results across classes, even though WRN-50 has the best AUROC for 3 UUCs in Vaihingen and 2 UUCs in Potsdam. When the UUC is Low Vegetation (\textit{2}) or High Vegetation (\textit{3}) almost all models and backbones have difficulties in identifying the OOD pixels. One possible reason for this poor performance is their high intraclass variability, hampering the characterization of OOD pixels.

For the Houston dataset, OpenIPCS with DN-121 achieved the best performance as shown in Table~\ref{tab:grss_results}. The proposed method was not able to outperform the state-of-the-art results given by OpenIPCS with DN-121. We hypothesize that this might be linked to the size of the training set on Houston, which is considerably smaller than Vaihingen and Potsdam.


\section{Conclusion}
\label{sec:conclusion}

\methodname{} outperformed, in all tested scenarios, the other models using the same closed set backbone. For Vaihingen and Potsdam datasets, \methodname{} established new state-of-the-art performance achieving the best average performance. 
Adapting other closed set backbones to \methodname{} is a challenge. The architecture uses spatial information from the Closed Set Encoder as input to the reconstruction decoder. Adapting a backbone with a different architecture will likely also imply changes in the CoReSeg architecture.

In our future work we consider: adapting other closed set backbones to work with the \methodname{} method; modifying the conditioning mechanism to better characterize each class; test on time series, sparsely labeled and not urban datasets.

\bibliographystyle{IEEEbib}
\bibliography{strings,refs}

\begin{thebibliography}{10}

\bibitem{krizhevsky2012imagenet}
Alex Krizhevsky, Ilya Sutskever, and Geoffrey~E Hinton,
\newblock ``{ImageNet Classification with Deep Convolutional Neural
  Networks},''
\newblock {\em NIPS}, vol. 25, 2012.

\bibitem{ren2015faster}
Shaoqing Ren, Kaiming He, Ross Girshick, and Jian Sun,
\newblock ``{Faster R-CNN: Towards Real-Time Object Detection with Region
  Proposal Networks},''
\newblock {\em NIPS}, vol. 28, 2015.

\bibitem{long2015fully}
Jonathan Long, Evan Shelhamer, and Trevor Darrell,
\newblock ``{Fully Convolutional Networks for Semantic Segmentation},''
\newblock in {\em CVPR}, 2015, pp. 3431--3440.

\bibitem{ronneberger2015}
Olaf Ronneberger, Philipp Fischer, and Thomas Brox,
\newblock ``{U-net: Convolutional Networks for Biomedical Image
  Segmentation},''
\newblock in {\em MICCAI}. Springer, 2015, pp. 234--241.

\bibitem{YUAN2021114417}
Xiaohui Yuan, Jianfang Shi, and Lichuan Gu,
\newblock ``{A Review of Deep Learning Methods for Semantic Segmentation of
  Remote Sensing Imagery},''
\newblock {\em Expert Systems with Applications}, vol. 169, pp. 114417, 2021.

\bibitem{geng2021}
Chuanxing Geng, Sheng-Jun Huang, and Songcan Chen,
\newblock ``{Recent Advances in Open Set Recognition: A Survey},''
\newblock {\em IEEE TPAMI}, vol. 43, no. 10, pp. 3614--3631, 2021.

\bibitem{silva2020}
Caio C.~V. da~Silva, Keiller Nogueira, Hugo~N. Oliveira, and Jefersson A.~dos
  Santos,
\newblock ``{Towards Open-Set Semantic Segmentation Of Aerial Images},''
\newblock in {\em Latin American GRSS/ISPRS Remote Sensing Conference
  (LAGIRS)}, 2020, pp. 16--21.

\bibitem{oliveira2021}
Hugo Oliveira, Caio Silva, Gabriel~L.S. Machado, Keiller Nogueira, and
  Jefersson~A. dos Santos,
\newblock ``{Fully Convolutional Open Set Segmentation},''
\newblock {\em Machine Learning}, pp. 1--52, 7 2021.

\bibitem{cui2020}
Zhiying Cui, Wu~Longshi, and Ruixuan Wang,
\newblock ``{Open Set Semantic Segmentation With Statistical Test And Adaptive
  Threshold},''
\newblock in {\em International Conference on Multimedia and Expo (ICME)},
  2020, pp. 1--6.

\bibitem{bendale2016}
Abhijit Bendale and Terrance~E. Boult,
\newblock ``{Towards Open Set Deep Networks},''
\newblock in {\em CVPR}, 2016, pp. 1563--1572.

\bibitem{perez2018film}
Ethan Perez, Florian Strub, Harm de~Vries, Vincent Dumoulin, and Aaron
  Courville,
\newblock ``{FiLM: Visual Reasoning with a General Conditioning Layer},''
\newblock {\em AAAI Conference on Artificial Intelligence}, vol. 32, no. 1,
  Apr. 2018.

\bibitem{oza_patel_2019}
P.~Oza and V.~M. Patel,
\newblock ``{C2AE: Class Conditioned Auto-Encoder for Open-Set Recognition},''
\newblock in {\em CVPR}, Los Alamitos, CA, USA, jun 2019, pp. 2302--2311, IEEE
  Computer Society.

\bibitem{GRSS}
Saurabh Prasad, Bertrand Le~Saux, Naoto Yokoya, and Ronny Hansch,
\newblock ``{2018 IEEE GRSS Data Fusion Challenge – Fusion of Multispectral
  LiDAR and Hyperspectral Data},'' 2020.

\end{thebibliography}

\end{document}